\def\year{2019}\relax
\documentclass[letterpaper]{article} 
\usepackage{aaai19}  
\usepackage{times}  
\usepackage{helvet} 
\usepackage{courier}  
\usepackage[hyphens]{url}  
\usepackage{graphicx} 
\usepackage{graphicx}  
\usepackage{floatrow}
\usepackage{adjustbox}
\usepackage{tikz}
\usetikzlibrary{arrows}

\title{Using Fractal Neural Networks to Play SimCity 1 and Conway's Game of Life at Variable Scales}
\author{Sam Earle\\
smearle93@gmail.com}
 
\begin{document}

\maketitle

\begin{abstract}
        We introduce gym-city, a Reinforcement Learning environment that uses \textit{SimCity 1}'s game engine to simulate an urban environment, wherein agents might seek to optimize one or a combination of any number of city-wide metrics, on gameboards of various sizes. 
        We focus on population, and analyze our agents' ability to generalize to larger map-sizes than those seen during training.
        The environment is interactive, allowing a human player to build alongside agents during training and inference, potentially influencing the course of their learning, or manually probing and evaluating their performance.
        To test our agents' ability to capture distance-agnostic relationships between elements of the gameboard, we design a minigame within the environment which is, by design, unsolvable at large enough scales given strictly local strategies. 
        Given the game engine's extensive use of Cellular Automata, we also train our agents to ``play'' Conway's Game of Life -- again optimizing for population -- and examine their behaviour at multiple scales. 
        To make our models compatible with variable-scale gameplay, we use Neural Networks with recursive weights and structure -- fractals to be truncated at different depths, dependent upon the size of the gameboard.

\end{abstract}

\section{Motivation}

Recent work in Reinforcement Learning has led to human- and superhuman-level performance in a variety of strategy games, in which players take actions on a game board of discrete tiles (e.g., Starcraft \cite{vinyals2019alphastar}, and Go \cite{silver2016alphagozero}).

However, the resultant agents' ability to generalize to adjacent tasks is not well established.
We consider agents that observe the entire board at each step, and test their ability to generalize by increasing its size.
On this larger board, local relationships between structures on the map remain the same, but the ways in which these might combine to form global dynamics have been altered with the size of the board.

We focus on agents that observe the entire board because we're interested in problems and solutions with a global component, where the entire board might come to bear on any given move.
The decision to take any particular action, then, should have the potential to be affected by the observed state of any tile.
We thus provide the entire board anew at each step, and in turn allow the agent to act on any tile.

Our work also finds motivation in the field of urban planning, where cities have increasingly come to be modeled as fractals \cite{batty1994fractalcities}.
We subscribe to this point of view, and ask if urban simulations can be used to generate such fractal urban forms.
In particular, we train agents using Reinforcement Learning to maximize populationa in urban simulations, in an effort to simulate urban growth.
Given that we expect the products of such a simulation to be of a fractal nature, we expect them to demonstrate structural similarity at various scales.

\section{Micropolis}
\begin{figure*}[t!]
    \includegraphics[width=500pt]{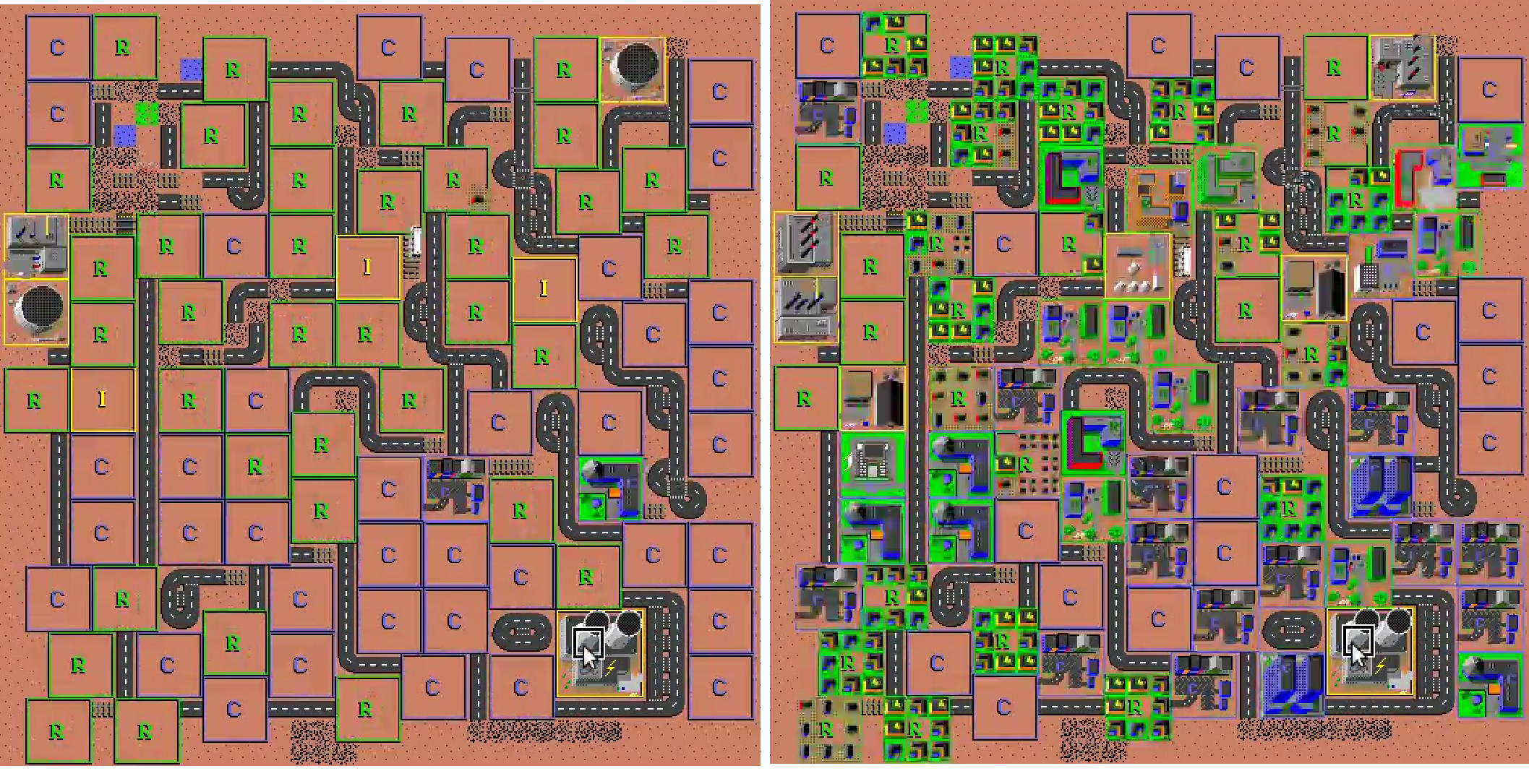}
	\caption{The result of a fractal neural network's deterministic action on a $32\times 32$ map, after training on a $16 \times 16$ map. A human player intervened minimally during the episode, deleting power plants to encourage exploration. The city fluctuates between an abandoned state (left) and a populous one (right).}
	\label{fig:simcity_w32.jpeg}
\end{figure*}
We use an open-source clone of SimCity 1 -- Micropolis -- to serve as a Reinforcement Learning Environment. Micropolis is a city-simulation game, in which the player starts with a procedurally-generated landmass, and is invited to populate it with urban structures. 
The map is a square grid of discrete tiles;
on each tile, the player can place power plants (nuclear or coal), power lines, zones (residential, commercial or industrial), services (fire stations or police departments), road, rail, parks, stadiums and airports.

Zones can only support development if they are connected to power, which emanates from power plants, and is conducted through adjacent zones, buildings and power lines.
The amount and type of development that arises depends on a number of factors, such as whether the zone is adjacent to road or rail, whether it is adjacent to a park or water (this can be seen to affect the property value of incoming development), the city-wide tax rate, crime, pollution, and demand (a measure of how much interest there is from outside the city in each type of development, which is in turn affected by the ratio of development types inside the city).

The player begins with a fixed sum of money, dependent upon a difficulty setting, which they can spend on construction and zoning, and ultimately make back in tax revenue. In addition to building and zoning, the player can control the city-wide tax rate and service budgets. 

The simulation depends largely on cellular automata \cite{wells2011new}. Besides being connected to the power grid -- which is easy to implement as a cellular automaton -- a residential zone will populate according to rules about what surrounds it.
When we place a single tile of road (even road nonadjacent to any other road, zones, or structures) next to a residential zone, the previous low-rise development will almost always turn into high-rise, and adding an adjacent park will likewise almost always cause the development to transition into the `high-income' version of itself, while adjacent or nearby coal power or polluting industry stifles development.
Traffic may flow between any two zones (populated, powered or not) connected by a path of adjacent road tiles, and often spreads indiscriminately to flow along any such path.

Incidentally, cellular automata have been studied extensively as a means of simulating urban growth; see \cite{aburas2016urbangrowth} for a review, and \cite{white1993automata} in particular for their use of cellular automata in generating fractal forms.

\subsection{Implementation and Interactivity}

SimCity 1 was open-sourced in 2008, as part of the One Laptop Per phild Program.
Its developer, Don Hopkins, refactored the code, adding a GUI written in Python.
The game engine, in C++, is interfaced with python via a fixed set of shared functions using Swift.
These functions can be called in Python to initialize, call, and query the game engine, and our environment uses them to affect and observe the game map.

In addition, changes can be made to the map by a human player, via this GUI, if it is rendered during either training or inference. 
Not only will the agent observe the new map-state at the next possible opportunity, but, during training, it will treat the player's actions as its own.
Implementation-wise, this means that when the player interacts with the GUI, building something on a particular tile, the build is queued, for the agent to execute at the next available opportunity, instead of whichever action had been randomly selected from its action distribution.
As a result, the bot can, in theory, learn to (dis)favour the types of action taken by the player, depending on their impact on reward, just as it would its own.

During testing of this feature, for example, the agents being trained would learn quite early on to place a single nuclear power plant amid a field of connected residential zones, resulting in fluctuating swaths of low-density development. 
Normally, agents would spend considerable time at this local optimum, but when a human player intervened in real time during training, relentlessly placing roads next to such residential zones and thus inviting stable, high density development, agents appeared to catch on to this behaviour, escaping the local optimum more rapidly than their peers.

During inference, the human player can manually probe an agent's learned policy in real time.
Often, when acting on larger maps than were trained upon, agents will build only on small subsections of the map, or fill the entire map but leave large subsections disconnected from power, or execute some otherwise quasi-effective urban plan.
We find that if a human player attacks the city's power plants, bulldozing them or replacing them with other structures, the agent is likely to improve upon its original city design -- laying new zones, zone-adjacent roads, and connecting structures to the power grid, for example -- before installing a new power plant and stabilizing once again.
We can thus sometimes coax out the scale-invariant side of an agent's behavioural policy by using a human adversary to incite chaos on the gameboard, forcing the agent to explore beyond local optima.

It can be argued that, in such an open-ended game as Micropolis, the goal of human play is to understand and master the principles underlying the simulation \cite{FM660}. 
When we extend the existing game with an RL agent, our goal shifts, as we navigate through its often intractable pathologies -- refractions of the game's design through the agent's neural architecture. 

\section{Experiment Design}

\subsection{Micropolis}

We give the agent a 2D ``image'' of the board, in which pixels correspond to tiles, and channels correspond to tilestates.
We include three additional channels of local information in our representation which correspond to population density (so our agent can ``see'' when development has arisen in a particular zone), traffic density, and whether or not a tile is powered.
Finally we include channels of global information, which take the same value at all tiles; these are overall population, and residential, commercial, and industrial demand.
At each step, the agent observes the gameboard, and outputs a 2D image of the same dimensions, whose channels correspond to tile-specific actions.
The agent may build any of the structures available to the player, in addition to land and water tiles.
In this action space, the agent specifies a single build to be carried out before the following step.

In our experiments, the agent is rewarded at each step by the city's population, plus a bonus for the number of distinct types of populated zone in the city. 
A step through the environment corresponds to 100 ticks in the game engine: this maximizes the positive feedback an agent is likely to receive from population-inducing tile configurations (for example, the placement of a residential zone next to a powered tile), especially early in training, when the agent is likely to overwrite them with other builds in fewer turns than it would take for population to arise on the zone tile at slower game speeds.

We give our agent virtually unlimited funds, and no option to change the default tax rate and service budgets, focusing instead on its ability to place a coherent set of structures on the gameboard, and noting that it is possible to develop populous cities without changing these default values, and still a difficult, combinatorially explosive task to do so optimally with infinite funds.
We note also that a limit on funds would likely only punish the agent unduly during early training, when it is building overzealously and largely at random; moreover, most populous cities the agent ultimately learns to design receive a positive monthly income as a result of taxation. 

However, constraining the budget could provide an interesting avenue for future work.
We might, for example, add a new strategic element to the game by starting the agent with very limited funds, forcing it to start small in its city-planning and wait for taxes to come in before continuing construction.

We limit the agent to a $16\times16$--tile patch of empty map during most of our experiments. This is the smallest size at which the agent begins to experience reward, and thus learn, relatively quickly: any smaller and the agent will too often be overwriting its previous builds for it to stumble upon a chance adjacent power-and-residential pairing. 
By keeping the map small, we decrease the amount of time and compute needed to complete a training session.

\subsection{Power Puzzle}

\begin{figure}
    \includegraphics[width=150pt]{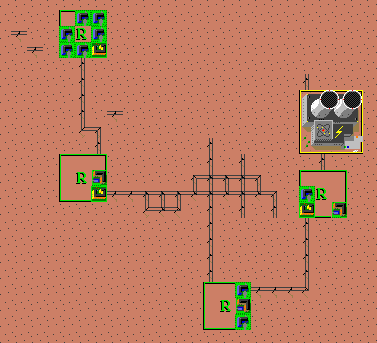}
    \caption{Deterministic performance of a Fractal Network on the power puzzle, on a map of width 20.}
    \label{fig:powpuz_w20}
\end{figure}

Given that Micropolis' logic is so inherently local, and that we supply certain global information at the local level in our experiment design, we might expect agents to develop highly local strategies.
We test an agent's ability to make connections between distant elements of the gameboard by designing a simple mini-game
(Figure \ref{fig:powpuz_w20}).
At the beginning of every episode of gameplay, between 1 and 5 residential zones are placed randomly on the map, followed by a nuclear power plant.
The agent's action-space is restricted to one channel; it may only build power wires, and cannot overwrite existing structures (the map is unchanged if it elects to build on such a tile). 

As before, we reward the agent, at each step, by the population on the gameboard.
Though it will be disparate and unstable in isolated residential zones, it is guaranteed to arise once they are powered, so that the agent is rewarded for each zone it manages to connect to the power grid.
In sum, we can expect the agent to find the the minimum spanning tree over nodes of the random power network: beginning from the power plant, and repeatedly connecting the nearest unconnected node.
In this mini-game, agents with local strategies will at best be able to branch out blindly from power sources to sinks, while an agent with the appropriate global strategy would immediately draw the shortest possible path between a source and a sink, even if these are far removed on the gameboard.

Incidentally, this mini-game can be solved by a Cellular Automaton, which can itself be implemented as a hand-weighted, recursive Neural Network. 
Unlike the networks we train with backpropogation, however, the only such implementation we have found demands a novel activation function.
Still, this coincidence helps motivate our use of fractal networks, which -- after the weight-sharing introduced below -- contain subnetworks otherwise identical in structure to our implementation of the CA; and could thus learn to implement something similar, perhaps using spare activation channels to work around the simplicity of the imposed 'ReLU' activation function.

\subsection{Game of Life}

Given Micropolis' heavy reliance on Cellular Automata, and the more general use of CA as models of urban life; and the fact that -- using Micropolis as an environment -- CA can be handcrafted to play certain simple, urban-planning mini-games; we also develop an environment that consists of a single, simple CA -- namely, Conway's Game of Life.
In our adaptation of GoL, the gameboard is populated randomly; each tile (or ``cell'') is ``alive'' with $20\%$ chance at the beginning of each episode.
The agent observes the board directly as a $1-$channel image, chooses one cell to bring to life, lets the automaton simulate through one tick, and repeats.
We reward for population, measured as the number of living cells on the gameboard, at each step.

Hypothetically, insofar as a successful agent merely adds to a non-empty and stable configuration without itself causing instability, we expect a relatively local strategy to suffice.
Insofar as the agent learns to stabilize chaotic gameboards, or, even moreso, as it learns to induce chaotic but consistently populous gameboards, we expect it to apply a more global strategy, wherein it must anticipate the dissipation of causal relations across the gameboard a few steps in advance to quell or control chaos.

\subsection{Network Architectures}

Because our agents observe the entire gameboard, whose dimensions need not be fixed in advance, our networks should accept inputs of various sizes.
We thus opt for networks made up of convolutional layers, which do exactly this.

Our simplest model is based on the ``FullyConv'' model in \cite{vinyals2017starcraft}, and consists of two distinct convolutions followed by ReLU activations, of $5 \times 5$ and $3 \times 3$ kernel size respectively. 
This is followed by a $1 \times 1$ convolution to produce the action distribution, and, to produce the value prediction, a dense hidden layer with 256 neurons, followed by a tanh activation, and a dense layer with scalar output.

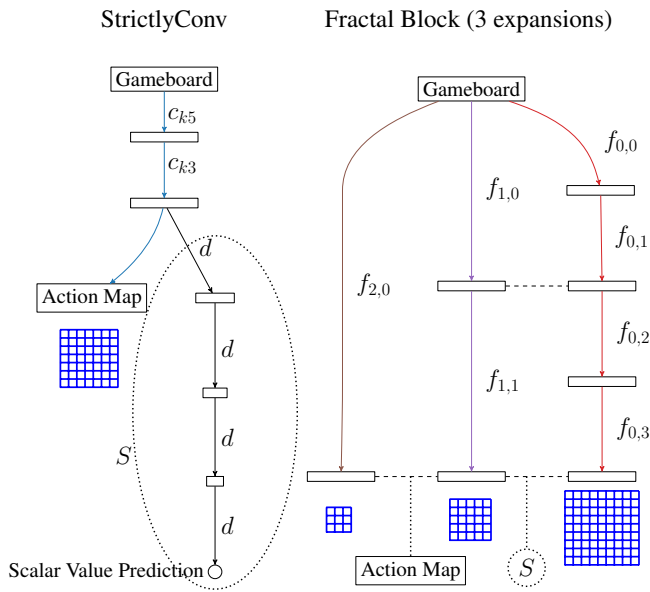
\begin{figure}[t!]
\begin{adjustbox}{width=250pt}
\begin{tikzpicture}[>=latex',line join=bevel,]
\draw (100bp, 400bp) node {\huge StrictlyConv};
\node (A) at (100.0bp,360bp) [draw,rectangle] {\LARGE Gameboard};
\node (B) at (100.0bp,320bp) [draw,rectangle] {\hspace*{40pt}$$};
  \node (C) at (100.0bp,274.5bp) [draw,rectangle] {\hspace*{40pt}$$};
  \node (D) at (50.0bp,209.0bp) [draw,rectangle] {\LARGE Action Map};
  \node (E) at (135.0bp,209.0bp) [draw,rectangle] {\hspace*{20pt}};
  \node (F) at (135.0bp,143.5bp) [draw,rectangle] {\hspace*{10pt}};
  \node (G) at (135.0bp,82.5bp) [draw,rectangle] {\hspace*{5pt}$$};
  \draw[dotted, line width=1] (135bp, 130bp) ellipse (2cm and 4.3cm);
  \node (H) at (135.0bp,20.0bp) [draw,circle] {\huge$$};
  \definecolor{strokecolor}{rgb}{0.12,0.47,0.71};
  \draw [strokecolor,-stealth'] (A) -> (B);
  \definecolor{strokecol}{rgb}{0.0,0.0,0.0};
  \pgfsetstrokecolor{strokecol}
  \draw (112.5bp,335bp) node {\huge $c_{k5}$};
  \draw (94.0bp,384.83bp) node {$$};
  \draw (94.0bp,344.21bp) node {$$};
  \definecolor{strokecolor}{rgb}{0.12,0.47,0.71};
  \draw [strokecolor,-stealth'] (B) -> (C);
  \draw (112.5bp,300.5bp) node {\huge $c_{k3}$};
  \draw (94.0bp,324.93bp) node {$$};
  \draw (94.0bp,284.28bp) node {$$};
  \definecolor{strokecolor}{rgb}{0.12,0.47,0.71};
  \draw [strokecolor,-stealth'] (C) ..controls (97.155bp,262.85bp) and (92.141bp,242.33bp)  .. (D);
  \definecolor{strokecolor}{rgb}{0,0,0};
  \draw [strokecolor,-stealth'] (C) ..controls (106.77bp,261.83bp) and (120.07bp,236.94bp)  .. (E);
  \draw (128.5bp,244.0bp) node {\huge $d$};
  \draw (95.882bp,264.98bp) node {$$};
  \draw (127.11bp,218.54bp) node {$$};
  \draw [strokecolor,-stealth'] (E) ..controls (135.0bp,196.59bp) and (135.0bp,172.88bp)  .. (F);
  \draw (143.5bp,174.0bp) node {\huge $d$};
  \draw (129.0bp,199.48bp) node {$$};
  \draw (129.0bp,153.04bp) node {$$};
  \draw [strokecolor,-stealth'] (F) ..controls (135.0bp,131.46bp) and (135.0bp,110.28bp)  .. (G);
  \draw (143.5bp,113.0bp) node {\huge $d$};
  \draw (129.0bp,133.87bp) node {$$};
  \draw (129.0bp,92.038bp) node {$$};
  \draw [strokecolor,-stealth'] (G) ..controls (135.0bp,70.483bp) and (135.0bp,50.078bp)  .. (H);
  \draw (143.5bp,52.0bp) node {\huge $d$};
  \draw (129.0bp,72.778bp) node {$$};
  \draw (129.0bp,31.325bp) node {$$};
  \draw (72bp,100.0bp) node {\huge $S$};
  \draw (60bp,21.0bp) node {\LARGE Scalar Value Prediction};
  \draw [step=0.2,blue, very thick] (.99,5.19) grid (2.4,6.6);

\end{tikzpicture}
\begin{tikzpicture}[>=latex',line join=bevel,]
	\draw (115bp, 392bp) node {\huge Fractal Block (3 expansions)};
\begin{scope}
  \pgfsetstrokecolor{black}
  \definecolor{strokecol}{rgb}{1.0,1.0,1.0};
  \pgfsetstrokecolor{strokecol}
  \definecolor{fillcol}{rgb}{1.0,1.0,1.0};
  \pgfsetfillcolor{fillcol}
  \filldraw (0.0bp,0.0bp) -- (0.0bp,356.94bp) -- (250.0bp,356.94bp) -- (250.0bp,0.0bp) -- cycle;
\end{scope}
\begin{scope}
  \pgfsetstrokecolor{black}
  \definecolor{strokecol}{rgb}{1.0,1.0,1.0};
  \pgfsetstrokecolor{strokecol}
  \definecolor{fillcol}{rgb}{1.0,1.0,1.0};
  \pgfsetfillcolor{fillcol}
  \filldraw (0.0bp,0.0bp) -- (0.0bp,356.94bp) -- (250.0bp,356.94bp) -- (250.0bp,0.0bp) -- cycle;
\end{scope}
  \node (1_1) at (117.0bp,209.44bp) [draw,rectangle] {\hspace*{40pt}$$};
  \node (2_2) at (207.0bp,209.44bp) [draw,rectangle] {\hspace*{40pt}$$};
  \node (1_2) at (117.0bp,77.941bp) [draw,rectangle] {\hspace*{40pt}$$};
  \node (2_4) at (207.0bp,77.941bp) [draw,rectangle] {\hspace*{40pt}$$};
  \node (0_1) at (27.0bp,77.941bp) [draw,rectangle] {\hspace*{40pt}$$};
  \node (A) at (117.0bp,345.44bp) [draw,rectangle] {\LARGE Gameboard};
  \node (2_1) at (206.0bp,275.44bp) [draw,rectangle] {\hspace*{40pt}$$};
  \node (B) at (75.0bp,13bp) [draw,rectangle] {\LARGE Action Map};
  \node (E) at (155.0bp,15bp) [draw,circle,dotted, line width=1] {\huge $S$};
  \node (2_3) at (207.0bp,143.44bp) [draw,rectangle] {\hspace*{40pt}$$};
  \draw [dashed] (1_1) ..controls (156.03bp,209.44bp) and (168.0bp,209.44bp)  .. (2_2);
  \draw [dashed] (1_2) ..controls (156.03bp,77.941bp) and (168.0bp,77.941bp)  .. (2_4);
  \draw [dashed] (0_1) ..controls (66.034bp,77.941bp) and (77.997bp,77.941bp)  .. (1_2);
  \definecolor{strokecolor}{rgb}{0.55,0.34,0.29};
  \draw [strokecolor,-stealth'] (A) ..controls (57.346bp,323.13bp) and (28.0bp,304.34bp)  .. (28.0bp,275.44bp) .. controls (28.0bp,275.44bp) and (28.0bp,275.44bp)  .. (28.0bp,143.44bp) .. controls (28.0bp,127.08bp) and (27.684bp,108.42bp)  .. (0_1);
  \definecolor{strokecol}{rgb}{0.0,0.0,0.0};
  \pgfsetstrokecolor{strokecol}
  \draw (49.5bp,209.44bp) node {\huge $f_{2,0}$};
  \definecolor{strokecolor}{rgb}{0.58,0.4,0.74};
  \draw [strokecolor,-stealth'] (A) ..controls (117.0bp,309.9bp) and (117.0bp,254.88bp)  .. (1_1);
  \draw (138.5bp,275.44bp) node {\huge $f_{1,0}$};
  \definecolor{strokecolor}{rgb}{0.84,0.15,0.16};
  \draw [strokecolor,-stealth'] (A) ..controls (190bp,321.56bp) and (200bp,302.07bp)  .. (2_1);
  \draw (220bp,308.44bp) node {\huge $f_{0,0}$};
  \draw [dotted, line width = 1] (75bp,78bp) -- (B);
  \draw [dotted, line width = 1] (155bp,78bp) -- (E);
  \definecolor{strokecolor}{rgb}{0.58,0.4,0.74};
  \draw [strokecolor,-stealth'] (1_1) ..controls (117.0bp,182.3bp) and (117.0bp,124.56bp)  .. (1_2);
  \draw (138.5bp,143.44bp) node {\huge $f_{1,1}$};
  \definecolor{strokecolor}{rgb}{0.84,0.15,0.16};
  \draw [strokecolor,-stealth'] (2_1) ..controls (206.26bp,258.51bp) and (206.53bp,240.27bp)  .. (2_2);
  \draw (227.5bp,242.44bp) node {\huge $f_{0,1}$};
  \definecolor{strokecolor}{rgb}{0.84,0.15,0.16};
  \draw [strokecolor,-stealth'] (2_2) ..controls (207.0bp,192.51bp) and (207.0bp,174.27bp)  .. (2_3);
  \draw (228.5bp,176.44bp) node {\huge $f_{0,2}$};
  \definecolor{strokecolor}{rgb}{0.84,0.15,0.16};
  \draw [strokecolor,-stealth'] (2_3) ..controls (207.0bp,126.73bp) and (207.0bp,108.89bp)  .. (2_4);
  \draw (228.5bp,110.44bp) node {\huge $f_{0,3}$};
\newcommand*{\xMin}{1}%
\newcommand*{\xMax}{9}%
\newcommand*{\yMin}{4}%
\newcommand*{\yMax}{12}%

\draw [step=0.2,blue, very thick] (0.59,1.39) grid (1.2,2);
\newcommand*{\xMn}{16}%
\newcommand*{\xMx}{24}%
\newcommand*{\yMn}{4}%
\newcommand*{\yMx}{12}%

    \draw [step=0.2,blue, very thick] (3.59,1.19) grid (4.6,2.2);
\newcommand*{\xMnn}{32}%
\newcommand*{\xMxx}{40}%
\newcommand*{\yMnn}{4}%
\newcommand*{\yMxx}{12}%

\draw [step=0.2,blue, very thick] (6.39,0.59) grid (8.2,2.4);
\end{tikzpicture}
\end{adjustbox}
\caption{Using fractal expansion to increase receptive-field size beyond that of baseline models. $c_{k}$ are convolutions of various kernel widths, $d$ are strided convolutions, $f_{ij}$ are convolutions of kernel width 3, and $S$ corresponds to the value-prediction subnetwork that appears in both models. With intra-column weight-sharing, for every $i$, and any $j$, $j'$; $f_{ij} = f_{ij'}$. With inter-column weight-sharing, for any $i$, $i'$, and $j$, $j'$; $f_{ij} = f_{i'j'}$. Receptive fields of output neurons are shown in blue. Dashed lines represent averaging.}
	\label{fig:net_graph}
\end{figure}

But the dense hidden layer in the value-prediction subnetwork must be initialized with the input dimesions known in advance, so, in ``StrictlyConv,'' we replace it with a single convolution with a stride of 2, repeatedly applied as many times as necessary to yield a scalar value prediction (Figure \ref{fig:net_graph}). This allows for our input dimensions to be at least any square with side-length corresponding to some integer power of 2, though other rectangles are possible in practice, with the borders of certain activations going ignored by the subsequent convolutional layers. 
Both models have 32 channels in all convolutional layers.

We thus have a simple baseline capable of operating at multiple scales. However, neurons in its action ouput have limited receptive fields, which prevent the model from acting immediately on long distance relations between structures on the gameboard.
Even if we were to make StrictlyConv recurrent, it would take several turns before being able to associate structures on the map which are not captured by the agent's receptive field, diffusing them spatially through its 2D memory.
Without recurrence, our more narrow-sighted agents are confined to using the board as external memory; for example, by building paths of wire from a source or sink, which (given our representation of the board) carry information about the connected zone's powered-ness across the board itself. 

We could increase the receptive field of such a model, without adding new parameters, simply by repeating the $3 \times 3$ convolution in the body of the network, so that it gets applied multiple times in sequence.
However, this can make the network prohibitively deep, causing instability during training.

To stabilize training, we use skip connections, allowing shallower paths from input to output to run in parallel with deeper ones. 
To structure these skip connections, we follow the example of Fractal Neural Networks \cite{larsson2018fractalnet}, whose network architectures are given by the repeated application of a fractal expansion rule.
In the body of our network, we use one fractal ``block,'' the result of 5 fractal expansions, and 32 channels in the hidden 2D activations.

To build a fractal block, we begin with a single convolution.
With each fractal expansion, the structure of the network is obtained as follows: two copies of the existing network are stacked, and a single convolution is added as a skip connection, which receives the same input as the stack, and whose output is averaged with the stack's, to produce the overall block's output. 
After 5 expansions, the convolution with which we began has produced $2^{5-1}=16$ vertical copies of itself, and we can extract from the overall block a 16-layer-deep subnetwork consisting only of these convolutions in sequence, which we'll refer to as the $0^{th}$ ``column'' of our network.
The $0^{th}$ column has a receptive field of $33 \times 33$, so that, at its output, an action in one corner of the board can be affected by an observation in the opposite corner.
The $1^{st}$ column of our network has as its seed the convolution introduced as skip connection in the first expansion, leading to a column of depth $2^{4-1}=8$.
The other columns, in order, have depth 4, 2, and 1.

We train a single block of 5 columns.
Unlike the authors, we try sharing weight layers between the copies of subfractals created during expansion, so that each column taken individually corresponds to the repeated application of a single convolution -- we refer to this variation as intra-column weight-sharing.
With intra-column weight-sharing, each column can be conceived of as a continuous-valued Cellular Automaton \cite{griffeath2003new}, which 'ticks' as many times as there are layers in the column, during each pass through the network.

We also try inter-column weight sharing, wherein the entire fractal block uses one unique layer of convolutional weights, so that each column taken individually repeats the same convolution a different number of times.
Under this constraint, each column corresponds to a variable number of ticks through the same continuous-valued CA.

We train Fractal Networks with local and global drop-path as outlined in the original paper, which puts pressure on individual subnetworks -- especially columns -- to learn the task independently (so long as these subnetworks span from input to output in the original network and have no other leaves). 
Our agents are trained using the A2C algorithm given in \cite{mnih2016a2c}. Like the StrictlyConv network, the Fractal Network's body is followed by a $1 \times 1$ convolution to produce the action distribution, and a repeated, strided convolution to produce the scalar value prediction.

\begin{figure*}
    \includegraphics[width=500pt]{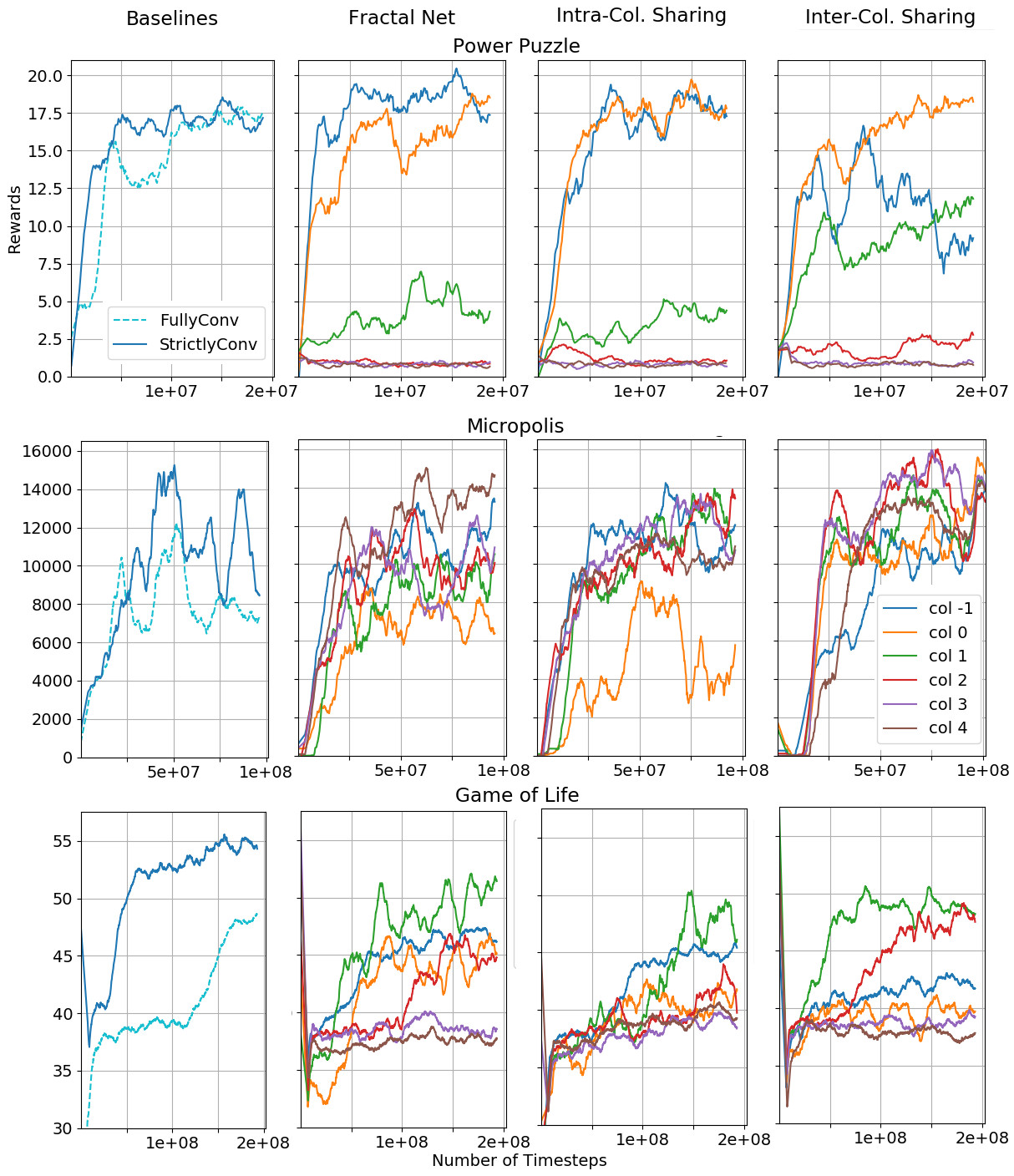}
       \caption{ Baselines and Fractal Networks, evaluated over the course of training. 'col -1' refers to the whole network.}
    \label{fig:evals}
\end{figure*}
\section{Results}

All experiments were executed using 96 concurrent environments to train the agent, with an Intel i7 CPU and one Nvidia GTX 1080 Ti GPU; each run completed within at least approximately 24 hours. 
The Power Puzzle was fastest, at only 20M frames, taking several hours, followed by the Game of Life, taking roughly half a day, with Micropolis being the most time-consuming.
There is likely room to optimize GoL, by instantiating the environment itself (a CA) as a hand-weighted Convolutional Neural Network, and offloading it to the GPU.

\subsection{Power Puzzle}

In the Micropolis Power Puzzle minigame, StrictlyConv matches FullyConv's performance, and does better earlier on in training .
Since both of these models have receptive fields of $7 \times 7$, they will not be able to identify the shortest path between a pair of zones that are more than 7 tiles apart horizontally or vertically.
To deal with these cases, the model can at best learn to branch out blindly from existing nodes.

The fractal network, with its larger receptive fields, is able to surpass the performance of both of the baseline models.
Without weight sharing, the overall block dominates performance, followed by the deepest, then second-deepest columns, and then by the rest, which learn to do virtually nothing useful.
The fact that deeper columns exhibit superior performance is not surprising: their larger receptive fields allow them to act on long distant connections. 

What is surprising is the failure of shallower columns. Column 2, for example, has a $9 \times 9$ receptive field, larger than that of the FullyConv model, at $7 \times 7$, but column 2 learns to do nothing independently.
We interpret this as the result of a co-adaptation between these shallower columns and the rest of the network, which drop-path is unable to mitigate: the shortsightedness of shallower columns is just too detrimental to the performance of the overall network for them to be allowed to act alone.
Interestingly, by sharing weights within and between columns of the fractal, we can increase the autonomy of individual columns at a slight cost to overall performance.

During interactive evaluation, we note that placing one or two power-wires adjacent to a source can result in the agent continuing construction in this direction until it either gets close enough, and subsequently connects, to a sink, or hits the edge of the map.
Indeed, agents can be deceived, especially on larger gameboards, to reach out in the wrong direction (i.e., away from the nearest zone).

\subsection{Micropolis}

When maximizing population in Micropolis, StrictlyConv outperforms FullyConv considerably, but both are unstable, and do not maintain peak performance for very long.
The fractal networks, at their best, narrowly outperform the baseline models. 
Though the reward of each subnetwork still fluctuates over the course of training, rarely do all subnetworks experience a simultaneous dip in performance.

With weight sharing disabled, the deepest column of the fractal network performs the worst, and the shallowest, the best, on the Micropolis population game.
When intra-column weight sharing is enabled, the deepest column suffers drastically, while the others are more tightly-clustered.
With inter-column sharing enabled, the performance ceiling increases again, and the deepest column is much more successful.
This indicates that by training a single set of weights to perform a task as a shallow recursive network, we have facilitated training the same set of weights, on the same task, as a much deeper recursive network.

The agents learn to densely fill maps of various sizes with zones of all types (Figure \ref{fig:simcity_w32.jpeg}), with an aversion toward industry.
It connects these zones to power largely without wire, simply by always placing new zones next to existing, powered ones.
It clusters zones by type, with larger clusters often forming on larger maps, and provides most zones access to a road tile, by spreading small cul-de-sacs throughout the map.
The agent exploits the traffic simulation: its roads somehow largely avoid attracting any traffic -- which would create development-inhibiting pollution -- but still allow adjacent zones to support high-density development.
Our agents can support stable populations on small maps, but cannot properly manage demand at larger scales, resulting in maps whose population fluctuates violently.

\subsection{Game of Life}

In maximizing population in GoL, StrictlyConv vastly outperforms FullyConv.

A fractal block without weight-sharing falls short of StrictlyConv, but its 8-layer deep column outperforms FullyConv. Its 2-layer and 1-layer columns perform the worst on the task. 
This suggests that neurons in the action distribution of the network are best off with receptive fields approximately the size of the gameboard. 
The 8-layer Column 1 is also closest in terms of receptive-field size to our baseline models; perhaps this column is only held back by the others.
We would be keen to retrain the fractal models without drop-path, allowing the other columns to co-adapt with column 1 (rather than learning the task independently), letting the latter take the lead.

Enabling weight-sharing does further damage to these other columns, suggesting that the policy learned by column 1 cannot be so easily scaled to use differently-sized receptive fields. 

\begin{figure}
    \includegraphics[width=\linewidth]{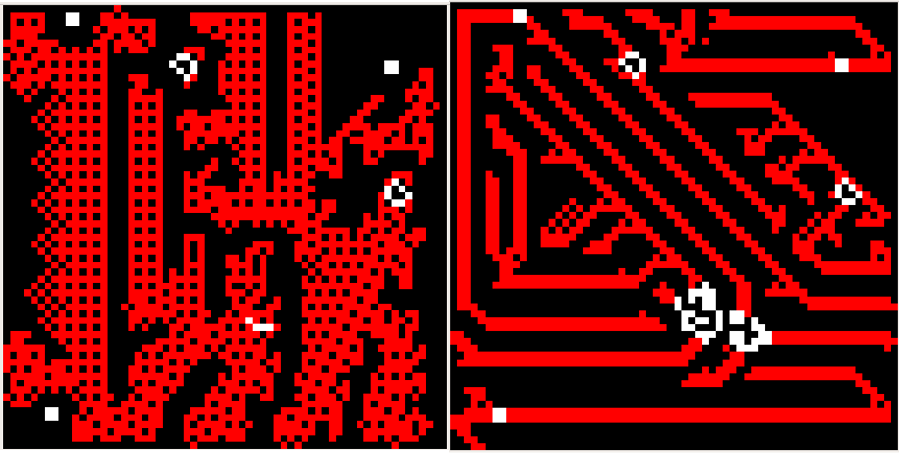}
    \caption{StrictlyConv (left) and Col. 1 of Fractal w/ intra- \& inter-col. sharing (right), at $64\times 64$. Cells placed by the agent are red.}
    \label{fig:GoL_builds}
\end{figure}

Behaviourally speaking, our agents tend to work around existing chaos, branching out from living cells to build space-filling paths which span the board in a few broad strokes before filling it out in more detail.
Indeed, the increased net activity on larger gameboards inhibits the agent's behaviour during inference; decreasing the initial population on these local gameboards, however, allows the agent to successfully scale up its strategy (Figure \ref{fig:GoL_builds}).

When a human player builds a living cell next to a structure built by the agent, it will often immediately build a new branch at this location.

\section{Conclusion}

StrictlyConv consistently matches or outperforms FullyConv; this may be because it has drastically fewer parameters (and the target value functions are simple in the environments studied here), because its recursive structure has a normalizing effect on the learned value function, or because it allows spatially-invariant processing of the map.
Setting all weights of the recursive value subnetwork to 1, for example, counts the total number of living cells in Game of Life, which is exactly the agent's reward.

Replacing a strictly convolutional network-body with a fractal block can provide narrow gains in the Micropolis and Power Puzzle environments.
By design, the Power Puzzle is the only task studied here that explicitly demands non-local gameplay strategies.
As evidenced by the baselines' performance on the task however, a clever enough 'blind' approach, branching out in various directions in search of proximal relationships, does quite well (at least on small maps).
Training on larger maps, with larger fractal blocks, and fewer zones to start, should more firmly evidence the advantage of large receptive fields in this task.

In Micropolis, the agent learns highly effective local strategies in its shallow columns.
When it is encouraged transfer these strategies to deeper subnetworks via-weight-sharing, overall performance increases, perhaps because the agent benefits from a more global approach that takes advantage of larger receptive fields.

Results in GoL reveal the primacy of local strategies, as agents learn how to place cells locally in order to build stable structures, which will result in neither death, nor new life.
We wonder if there is not a better strategy, however, in which the agent builds structures which result in a calculated bursting-forth of life, which may fill the board more quickly than the agent could on its own.
We could encourage such behaviour by decreasing the number of builds an agent makes during an episode, and/or alternating  between several agent builds, and several tick of the automata, allowing for the construction of distinct traveling/growing structures.

The fact that such drastic reductions to the number of parameters via weight-sharing in our fractal blocks often yields comparable performance is striking, though we cannot say for certain that this is not owing to the overall simplicity of the tasks at hand.

Conversely, the fact that one layer of weights can be repeatedly copied to create deep, trainable networks suggests an interesting avenue for Neuroevolution.
For instance, we might try, midway through training, replacing an individual layer of an existing network with a complex Fractal Block built up entirely of copies of the existing layer, which are then allowed to change independently, thus granting the network the capacity for more complex behaviour.

In general, using fractal expansion to evolve Neural Networks would allow models to directly repurpose or ``exapt'' existing structures (and the behaviours they instantiate), which could prove more effective than developing new ones from scratch.

\newpage
\bibliography{references}

\begin{thebibliography}{}

\bibitem[\protect\citeauthoryear{Aburas \bgroup et al\mbox.\egroup
  }{2016}]{aburas2016urbangrowth}
Aburas, M.~M.; Ho, Y.~M.; Ramli, M.~F.; and Ash’aari, Z.~H.
\newblock 2016.
\newblock The simulation and prediction of spatio-temporal urban growth trends
  using cellular automata models: A review.
\newblock {\em International Journal of Applied Earth Observation and
  Geoinformation} 52:380 -- 389.

\bibitem[\protect\citeauthoryear{Batty and
  Longley}{1994}]{batty1994fractalcities}
Batty, M., and Longley, P.
\newblock 1994.
\newblock {\em Fractal Cities: A Geometry of Form and Function}.
\newblock San Diego, CA, USA: Academic Press Professional, Inc.

\bibitem[\protect\citeauthoryear{Friedman}{1999}]{FM660}
Friedman, T.
\newblock 1999.
\newblock The semiotics of simcity.
\newblock {\em First Monday} 4(4).

\bibitem[\protect\citeauthoryear{Griffeath \bgroup et al\mbox.\egroup
  }{2003}]{griffeath2003new}
Griffeath, D.; Moore, C.; Moore, D.; and Institute, S.~F.
\newblock 2003.
\newblock {\em New Constructions in Cellular Automata}.
\newblock Proceedings volume in the Santa Fe Institute studies in the sciences
  of complexity. Oxford University Press.

\bibitem[\protect\citeauthoryear{Larsson, Maire, and
  Shakhnarovich}{2016}]{larsson2018fractalnet}
Larsson, G.; Maire, M.; and Shakhnarovich, G.
\newblock 2016.
\newblock Fractalnet: Ultra-deep neural networks without residuals.
\newblock {\em CoRR} abs/1605.07648.

\bibitem[\protect\citeauthoryear{{Mnih} \bgroup et al\mbox.\egroup
  }{2016}]{mnih2016a2c}
{Mnih}, V.; {Puigdom{\`e}nech Badia}, A.; {Mirza}, M.; {Graves}, A.;
  {Lillicrap}, T.~P.; {Harley}, T.; {Silver}, D.; and {Kavukcuoglu}, K.
\newblock 2016.
\newblock {Asynchronous Methods for Deep Reinforcement Learning}.
\newblock {\em arXiv e-prints}.

\bibitem[\protect\citeauthoryear{Silver \bgroup et al\mbox.\egroup
  }{2016}]{silver2016alphagozero}
Silver, D.; Huang, A.; Maddison, C.~J.; Guez, A.; Sifre, L.; van~den Driessche,
  G.; Schrittwieser, J.; Antonoglou, I.; Panneershelvam, V.; Lanctot, M.;
  Dieleman, S.; Grewe, D.; Nham, J.; Kalchbrenner, N.; Sutskever, I.;
  Lillicrap, T.; Leach, M.; Kavukcuoglu, K.; Graepel, T.; and Hassabis, D.
\newblock 2016.
\newblock Mastering the game of {Go} with deep neural networks and tree search.
\newblock {\em Nature} 529(7587):484--489.

\bibitem[\protect\citeauthoryear{Vinyals \bgroup et al\mbox.\egroup
  }{2017}]{vinyals2017starcraft}
Vinyals, O.; Ewalds, T.; Bartunov, S.; Georgiev, P.; Vezhnevets, A.~S.; Yeo,
  M.; Makhzani, A.; K{\"{u}}ttler, H.; Agapiou, J.; Schrittwieser, J.; Quan,
  J.; Gaffney, S.; Petersen, S.; Simonyan, K.; Schaul, T.; van Hasselt, H.;
  Silver, D.; Lillicrap, T.~P.; Calderone, K.; Keet, P.; Brunasso, A.;
  Lawrence, D.; Ekermo, A.; Repp, J.; and Tsing, R.
\newblock 2017.
\newblock Starcraft {II:} {A} new challenge for reinforcement learning.
\newblock {\em CoRR} abs/1708.04782.

\bibitem[\protect\citeauthoryear{Vinyals \bgroup et al\mbox.\egroup
  }{2019}]{vinyals2019alphastar}
Vinyals, O.; Babuschkin, I.; Chung, J.; Mathieu, M.; Jaderberg, M.; Czarnecki,
  W.~M.; Dudzik, A.; Huang, A.; Georgiev, P.; Powell, R.; Ewalds, T.; Horgan,
  D.; Kroiss, M.; Danihelka, I.; Agapiou, J.; Oh, J.; Dalibard, V.; Choi, D.;
  Sifre, L.; Sulsky, Y.; Vezhnevets, S.; Molloy, J.; Cai, T.; Budden, D.;
  Paine, T.; Gulcehre, C.; Wang, Z.; Pfaff, T.; Pohlen, T.; Wu, Y.; Yogatama,
  D.; Cohen, J.; McKinney, K.; Smith, O.; Schaul, T.; Lillicrap, T.; Apps, C.;
  Kavukcuoglu, K.; Hassabis, D.; and Silver, D.
\newblock 2019.
\newblock {AlphaStar: Mastering the Real-Time Strategy Game StarCraft II}.
\newblock
  \url{https://deepmind.com/blog/alphastar-mastering-real-time-strategy-game-starcraft-ii/}.

\bibitem[\protect\citeauthoryear{Wells}{2011}]{wells2011new}
Wells, M.
\newblock 2011.
\newblock New games of life: Cellular automata and subsurface discourses in
  simcity.

\bibitem[\protect\citeauthoryear{White and Engelen}{1993}]{white1993automata}
White, R., and Engelen, G.
\newblock 1993.
\newblock Cellular automata and fractal urban form: A cellular modelling
  approach to the evolution of urban land-use patterns.
\newblock {\em Environment and Planning A: Economy and Space} 25(8):1175--1199.

\end{thebibliography}
\end{document}